\documentclass[runningheads]{llncs}
\usepackage[T1]{fontenc}
\usepackage{graphicx}
\usepackage{booktabs}
\usepackage{multirow}
\usepackage{colortbl}
\usepackage{color}
\usepackage[pagebackref,breaklinks,colorlinks]{hyperref}

\usepackage{wrapfig}
\usepackage{orcidlink}
\usepackage{xr-hyper}
\usepackage{algorithm}
\usepackage{algpseudocode}
\usepackage{amsmath,bm}
\usepackage{bbm}
\usepackage{amsfonts}
\usepackage{caption}
\usepackage{subfig}

\usepackage{graphicx}

\definecolor{Gray}{gray}{0.9}
\definecolor{Better}{rgb}{0.18, 0.407, 0.266}
\definecolor{Worse}{rgb}{0.35, 0.35, 0.35}
\definecolor{drakgreen}{rgb}{0.38, 0.67, 0.38}
\definecolor{drakpurple}{rgb}{0.38, 0.27, 0.61}
\definecolor{granate}{rgb}{0.64, 0.16, 0.16}

\newcommand{\our}{\cellcolor{Gray}}
\newcommand{\best}[1]{\textbf{#1}}
\newcommand{\imp}[1]{$_{{\textbf{\textcolor{Better}{#1}}}}$}
\newcommand{\wor}[1]{$_{{\textbf{\textcolor{Worse}{#1}}}}$}
\newcommand{\nc}[1]{ \textcolor{granate}{#1}}

\usepackage[euler]{textgreek}
\usepackage{arydshln}
\usepackage{marvosym}

\newcommand{\ww}{\mathbf{w}}

\newcommand{\pp}{\mathbf{p}}                     
\newcommand{\xx}{\mathbf{x}}                     
\newcommand{\W}{\mathbf{W}}                     
\newcommand{\real}{\mathbb{R}}                   
\newcommand{\vv}{\mathbf{v}}                     
\newcommand{\ttt}{\mathbf{t}}                    
\newcommand{\temp}{\tau}                         
\newcommand{\domain}[1]{$\mathcal{D}_\text{#1}$} 

\begin{document}

\title{Full Conformal Adaptation of Medical Vision-Language Models} 
\titlerunning{Full Conformal Adaptation of Medical Vision-Language Models}
\author{Julio Silva-Rodríguez\inst{1}\textsuperscript{,\Letter} \and Leo Fillioux\inst{2} \and Paul-Henry Cournède\inst{2} \and Maria Vakalopoulou\inst{2} \and Stergios Christodoulidis\inst{2} \and Ismail {Ben Ayed}\inst{1} \and Jose Dolz\inst{1} }
\authorrunning{J.~Silva-Rodríguez et al.}
\institute{
~\inst{1}ÉTS Montréal \\
\Letter {\tt \small \email{julio-jose.silva-rodriguez@etsmtl.ca}} \\
~\inst{2}CentraleSupélec, Université Paris-Saclay
}
\index{Silva-Rodríguez, Julio}
\index{Fillioux, Leo}
\index{Cournède, Paul-Henry}
\index{Vakalopoulou, Maria}
\index{Christodoulidis, Stergios}
\index{Ben Ayed, Ismail}
\index{Dolz, Jose}

\maketitle         

\begin{abstract}

Vision-language models (VLMs) pre-trained at large scale have shown unprecedented transferability capabilities and are being progressively integrated into medical image analysis. Although its discriminative potential has been widely explored, its reliability aspect remains overlooked. This work investigates their behavior under the increasingly popular split conformal prediction (SCP) framework, which theoretically guarantees a given error level on output sets by leveraging a labeled calibration set. However, the zero-shot performance of VLMs is inherently limited, and common practice involves few-shot transfer learning pipelines, which cannot absorb the rigid exchangeability assumptions of SCP. To alleviate this issue, we propose \textit{full conformal adaptation}, a novel setting for jointly adapting and conformalizing pre-trained foundation models, which operates transductively over each test data point using a few-shot adaptation set. Moreover, we complement this framework with SS-Text, a novel training-free linear probe solver for VLMs that alleviates the computational cost of such a transductive approach. We provide comprehensive experiments using $3$ different modality-specialized medical VLMs and $9$ adaptation tasks. Our framework requires \textit{exactly} the same data as SCP, and provides consistent relative improvements of up to 27$\%$ on set efficiency while maintaining the same coverage guarantees. Code is available: \url{https://github.com/jusiro/FCA} .

\keywords{VLMs \and Transfer learning \and Conformal prediction}
  
\end{abstract}

\section{Introduction}
\label{sec:intro}

Large pre-trained vision-language models (VLMs), such as CLIP \cite{radford2021learning} or ALIGN \cite{jia2021scaling}, are becoming increasingly popular in a plethora of computer vision problems \cite{lp24,luo2023segclip,clap24}, exhibiting unprecedented transfer capabilities to downstream tasks. Nevertheless, as these models gain traction in safety-critical scenarios, such as healthcare \cite{shakeri2024few,MedCLIP,FLAIR,CONCH}, ensuring their reliability and safety is paramount to prevent potential failures and thus minimize risks to human life. Therefore, the safe deployment of VLMs in high-stakes tasks requires not only precise discriminative performance but also robust uncertainty estimates.

Conformal Prediction (CP) \cite{learning_ny_transduction,pmlr-v25-vovk12,vovk_trans_conf,vovk_book} is a machine learning framework that provides model agnostic, and \textit{distribution-free}, finite-sample validity guarantees for handling reliability, and which has recently been adopted in modern neural networks \cite{raps,lac,aps,kim2020predictive,xu2021conformal}. Specifically, CP defines a non-conformity score function to produce a finite prediction set for a given unseen data point, which includes the true label with a specified
confidence level. For example, in the medical context, CP can ensure that the correct diagnosis (e.g., presence of a given disease) is included in the predictive sets 95\% of the time.

Split conformal prediction (SCP) \cite{inductive_ci,vovk_book,lei2018distribution} is arguably the most popular CP approach for uncertainty quantification, enjoying widespread adoption. SCP relies on a \textit{labeled calibration subset} to find a threshold on the non-conformity score that ensures a desired marginal coverage level on test data, whose distribution must be, at least, \textit{exchangeable}. w.r.t. calibration. Despite its increasing popularity, the rigid assumptions of SCP limit its successful deployment for foundation models. Even though VLMs have shown promising zero-shot capabilities, they usually require an adaptation stage to perform properly, e.g., in low-prevalence concepts or under domain shifts \cite{radford2021learning,udandarao2024zeroshot}, where the few-shot paradigm is becoming popular \cite{clap24}. However, the problem becomes evident when, if the non-conformity scores are re-adjusted based on such limited calibration data, the covariate shift breaks the exchangeability principle w.r.t. unseen examples. This raises the following question: \textit{can we adapt modern VLMs to downstream tasks yet ensure coverage guarantees on new data without relying on multiple, data-expensive, labeled subsets?} To address this question, we revisit full conformal prediction (FCP) \cite{learning_ny_transduction,transductive_ci}, which trains the model in a transductive fashion to ensure exchangeable reliability measures. However, its usability in modern deep neural networks is unclear since FCP requires multiple model fits for each query sample — concretely, one for each label — which explains the broader adoption of SCP. In this work, we accommodate this transductive setting to overcome the limitations observed in prior literature. The main contributions are:
\begin{itemize}
    \item[$\circ$] We introduce \textit{full conformal adaptation} (FCA), a novel framework that operates transductive to adapt pre-trained VLMs and create conformalized outputs with validity guarantees, based on a few-shot adaptation subset.
    \item[$\circ$] Furthermore, we complement FCA with a \textit{training-free} linear probing solver, SS-Text, which reduces $150\times$ the computational burden of adapting VLMs compared to SoTA approaches while providing robust performance.
    \item[$\circ$] Comprehensive experiments on 9 public datasets covering 3 different medical image modalities demonstrate the proposed approach's superiority compared to SCP, and the potential of conformal prediction in medical image analysis.
\end{itemize}

\section{Related Work}
\label{sec:rw}

\noindent\textbf{\textit{Transfer learning in VLMs}.} VLMs exhibit outstanding general zero-shot capabilities \cite{radford2021learning}. However, this good performance vanishes when specific domains and concepts are represented with low frequency during pre-training \cite{udandarao2024zeroshot}. This limitation motivates its data-efficient, few-shot adaptation to novel classes \cite{zhou2022coop,gao2021clip,zhang2021tip,kgcoop23,clap24}. For example, Prompt Learning aims to optimize a set of learnable input tokens for each task \cite{zhou2022coop,kgcoop23}. Black-box Adapters, which directly operate on feature representations, have emerged as a more efficient alternative, e.g., see \cite{lp24,shakeri2024few} for a direct comparison. Indeed, advanced linear probing methods that combine visual and text knowledge \cite{yu2023task,lin2023crossmodal,clap24,lp24} currently yield the best results. For instance, CLAP \cite{clap24} follows a constrained optimization objective, whereas LP++~\cite{lp24} combines visual and text logits through trainable blending parameters. In contrast to prior literature, which mostly studies VLM adaptation from a discriminative standpoint, our work explores its overlooked reliability.

\noindent\textbf{\textit{Conformal prediction in vision}.} Current trends on CP for image classification leverage black-box predictors using split conformal prediction \cite{aps,raps,conftr,conftr_24,einbinder2022training}. These works are usually evaluated on general vision tasks; thus, their performance is yet to be explored in medical imaging. From the technical side, recent efforts are devoted to better non-conformity measures, which account for better adaptability and set efficiency. For example, LAC \cite{lac} creates predictive sets by directly using the raw class probabilities. Adaptive Prediction Sets (APS) \cite{aps} computes the non-conformity score by accumulating the sorted softmax values in descending order, and its regularized extension RAPS \cite{raps} integrates explicit penalties to enforce small sets. However, these works employ models trained on an independent and identically distributed (i.i.d.) dataset w.r.t. testing, an unrealistic scenario in the era of foundation models.

\section{Background}
\label{sec:background}

\subsection{Zero-shot models}
\label{ssec:zero-shot}

\noindent\textbf{\textit{Contrastive vision-language models}.} VLMs usually follow the CLIP \cite{radford2021learning} setting and are trained on large heterogeneous datasets to encode similar representations between paired image and text information. CLIP-alike models comprise a vision encoder, $f_\theta(\cdot)$, and a text encoder, $f_\phi(\cdot)$, which project data points into an $\ell_{2}$-normalized $D$-dimensional shared embedding space, yielding the corresponding visual, ${\vv} \in \real^{D \times 1}$, and text, ${\ttt} \in \real^{D \times 1}$, embeddings. These models provide strong representations, which can be transferred in a black-box manner \cite{gao2021clip,zhang2021tip,ouali2023black,yu2023task,lin2023crossmodal,clap24,lp24}. Given a pre-computed image feature and class-wise prototypes, $\W=(\ww_c)_{1 \leq c \leq C}$, with $\ww_c  \in \real^{D \times 1}$, and $C$ the number of target categories, probabilities can be computed as:
\begin{align}
\label{eq:probs}
    \phantom{,}p_c(\W)
    = \frac
    {\exp( \vv^\top \ww_{c} / \temp)}
    {\sum_{j=1}^{C} \exp( \vv^\top \ww_j / \temp)},
\end{align}
where $\temp$ is a temperature parameter learned during the pre-training, $\vv^\top \ww$ is the dot product, equivalent to cosine similarity, as vectors are $\ell_2$-normalized, and $\pp(\W)=(p_c(\W))_{1 \leq c \leq C}$ corresponds to the predicted probabilities vector.

\noindent\textbf{\textit{Zero-shot inference}.} Contrastive VLMs enable zero-shot predictions, i.e., no need to explicitly learn $\W$, by embedding a textual description for each label, so-called prompt. Thus, given a set of $J$ text prompts, $\{\{\ttt_{cj}\}_{j=1}^{J}\}_{c=1}^{C}$, a common practice is to obtain a zero-shot prototype for each target category by computing the average of the $\ell_2$-normalized text embeddings for each class, $\ttt_{c}=\frac{1}{J}\sum_{j=1}^{J}\ttt_{cj}$. These prototypes can be readily integrated in Eq. \ref{eq:probs}, such that $\W^*=(\ttt_c)_{1 \leq c \leq C}$.

\subsection{Conformal prediction}
\label{ssec:conformal}

\noindent\textbf{\textit{Preliminaries}.} Let us define a multi-class image classification task composed of image and label data pairs $(\xx,y)$, randomly sampled from a joint distribution $\mathcal{P}_{\mathcal{XY}}$. Also, we denote a black-box deep network, $\pi(\cdot)$, which outputs probability assignments for each category, $\pp=\pi(\xx)$, in a label space, $\mathcal{Y}=\{1, 2, ..., C\}$, and which is trained on a base subset of $R$ samples from $\mathcal{XY}$, \domain{train}$=\{(\xx_r,y_r)\}_{r=1}^{R}$.  

\noindent\textbf{\textit{Conformal prediction}.} CP \cite{vovk_book} aims to produce prediction sets containing the ground truth label with a user-specified probability. Formally, the goal is to construct a set-valued mapping $\mathcal{C}:\mathcal{X}\rightarrow 2^{C}$, such that:
\begin{align}
\label{eq:marginal}
    \mathcal{P}(Y\in \mathcal{C}(\xx)) \geq 1-\alpha,
\end{align}
where $\alpha \in (0, 1)$ denotes the desired error rate, and $\mathcal{C}(\xx) \subset \mathcal{Y}$ is the prediction set. Eq. \ref{eq:marginal} is known as the \textit{coverage guarantee} \cite{vovk_book}, and is \textit{marginal} over $\mathcal{XY}$. 

\noindent\textbf{\textit{Split conformal prediction}.} SCP \cite{inductive_ci} enables deploying coverage guarantees for any pre-trained predictor \cite{Lei2018}. This framework solely requires access to $N$ labeled calibration data points, \domain{cal}$=\{(\xx_i,y_i)\}_{i=1}^{N}$, and $M$ test data samples, \domain{test}$=\{(\xx_m)\}_{m=N+1}^{N+M}$, both drawn from i.i.d or exchangeable distributions \cite{vovk_book}. The SCP procedure is as follows: \textit{i}) \textit{First} a non-conformity measure $s_{i}=\mathcal{S}(\pi(\xx_i),y_i)$ is evaluated, where $s_{i}$ is a measure of deviation; \textit{ii}) \textit{Second}, the 1-$\alpha$ quantile of the non-conformity score is determined from calibration data, which will serve as a confidence threshold to satisfy a given coverage:
\begin{align}
\label{eq:threshold}
    \hat{s} = \text{inf}
    \biggl[
    s : 
    \frac{ |i\in \{1,...,N\}: s_{i} \leq s| } { N }
    \geq 
    \frac{ \lceil (N+1)(1-\alpha) \rceil } { N }  
    \biggr];
\end{align}
\textit{iii}) \textit{Third}, for each testing sample, the non-conformity scores are calculated, and the prediction set comprises labels whose non-conformity score falls within $\hat{s}$:
\begin{align}
\label{eq:inference}
    \mathcal{C}(\xx) = \{ y \in \mathcal{Y} : \mathcal{S}(\pi(\xx),y) \leq \hat{s} \}.
\end{align}

\section{Methods}
\label{sec:methods}

\subsection{Problem statement}
\label{ssec:problem}

\noindent\textbf{\textit{Pitfalls of SCP in VLMs}.} A relevant corpus of recent works is popularizing SCP in deep vision networks \cite{raps,conftr,ding2024class}. Nevertheless, prior art focuses on narrow scenarios where the black-box model has been intensively trained on a dataset that is in-distribution to calibration/testing. However, this scenario is unrealistic in the era of foundation models. Their zero-shot performance is limited to the pre-training concept frequency and therefore requires adaptation \cite{udandarao2024zeroshot}. This adaptation is usually carried out in the few-shot data regime \cite{radford2021learning,clap24} leveraging a small balanced support set of $N=K\cdot|\mathcal{Y}|$, typically with $K\leq16$. With such limited supervision, SCP is naturally constrained. If the model is adapted in the support set, then the new scores would not be exchangeable w.r.t. test data and, as shown in Table \ref{main_results}, the marginal coverage in Eq. \ref{eq:marginal} will be violated.  

\noindent\textbf{\textit{Transductive conformal prediction}.} To address the above limitations, we leverage well-established knowledge in transductive prediction, particularly \textit{full conformal prediction} \cite{learning_ny_transduction,transductive_ci,trans_conf_machine}. FCP provides a more flexible approach in which the model is trained transductively for each test data point $\xx_m$ jointly with the training set without requiring calibration data. The intuition is that the true category for $\xx_m$ lies within $\mathcal{Y}$. Thus, if the model  $\pi(\cdot)$ is trained for each possible label assignment, i.e., $\pi(\cdot)^y : y_m = y \in \mathcal{Y} $, then the score distribution obtained from the correct label will be exchangeable to training score distribution. Therefore, the label-specific non-conformity measure given the trained model, $s_{i}^y=\mathcal{S}(\pi_i^y(\xx),y_i)$, can be used to compute the $1-\alpha$ quantile from the training score distribution, $\hat{s}^y$, similarly to Eq. \ref{eq:threshold}, and output predictive sets: 
\begin{align}
\label{eq:sets_full}
    \mathcal{C}(\xx) = \{ y \in \mathcal{Y} : s^y \leq \hat{s}^y \},
\end{align}
whose output sets satisfy the coverage guarantee in Eq. \ref{eq:marginal} if fitting $\pi(\cdot)$ is invariant to permutations, and assuming exchangeable data distributions \cite{vovk_book}. 

Nevertheless, FCP has been typically discarded due to its expensive computations, which require $C$ model training for each test image, being of impractical deployment when using modern deep networks. Indeed, the FCP procedure involves fitting the whole network on each label $y \in \mathcal{Y}$ and each test point, i.e., $\mathcal{D}_{train}=\{(\xx_1,y_1),...,(\xx_r,y_r),...(\xx_{R},y_{R}), (\xx_{m},y)\}$.

\subsection{Full conformal adaptation}
\label{ssec:fca}

\begin{figure*}[h!]
    \begin{center}
        \begin{tabular}{ccc}

         \includegraphics[width=.33\linewidth]{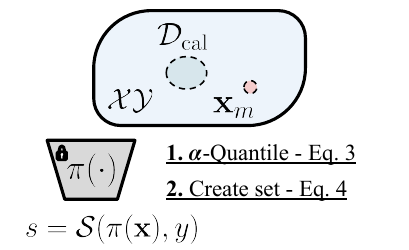} &
         \includegraphics[width=.33\linewidth]{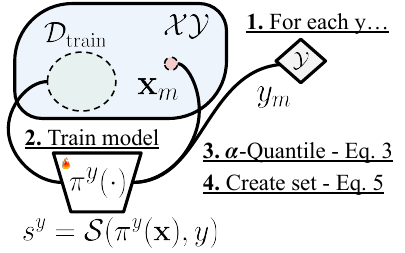} &
         \includegraphics[width=.33\linewidth]{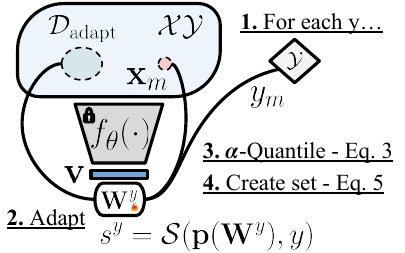} \\

         \textbf{(a) SCP} & \textbf{(b) FCP} & \textbf{(c) FCA (\textit{Ours})} \\

        \end{tabular}
        \caption{\textbf{Overview of conformal frameworks}: (a) split conformal prediction \cite{inductive_ci}, (b) Full conformal prediction \cite{learning_ny_transduction}, and (c) full conformal adaptation (\textit{Ours}).}
        \label{fig:overviews}
    \end{center}
\end{figure*}

There exist two limitations that prohibit FCP deployment in modern vision-language models: \textit{i}) accessing the pre-training dataset, and \textit{ii}) the computational burden of training largely parametrized networks for each test sample. In the following, we describe \textit{full conformal adaptation}. This novel framework tackles such issues and allows for deploying reliable yet precise solutions in modern vision-language models. An overview is presented in Fig. \ref{fig:overviews}, and we describe its key methodological components above:
\begin{itemize}
    \item[$\circ$] \textit{Using an adaptation set}. Instead of training the model from scratch, we propose to \textit{adapt} it using a small (few-shot) subset, similar to the one employed for calibration in SCP, i.e., \domain{adapt}$\equiv$\domain{cal}, thus omitting the pre-training data.
    \item[$\circ$] \textit{Transfer learning}. Instead of updating the whole network parameters, we propose operating over the black-box features extracted from the adaptation set, $\{(\vv_i,y_i)\}_{i=1}^{N}$, and the new test data point, $\vv_m$. Thus, model training in the transductive conformal setting can be reduced to finding the optimum linear probe for each explored label, $\W^y$, such that $s_{i}^y=\mathcal{S}(\pp_i(\W^y),y_i)$.
\end{itemize}

\noindent The proposed setting satisfies the coverage guarantees as FCP (see \cite{learning_ny_transduction,transductive_ci,trans_conf_machine}) but translates the exchangeability assumption to the feature space. Also, note FCA requires \textit{exactly} the same data sources as SCP, yet is more data efficient, as it allows both adaptation and conformalization. However, popular linear probing strategies might still be time-expensive — see Figure \ref{fig:accuracy_efficiency}(b). Motivated by this observation, we propose a novel solver for finding $\W^y$ more efficiently next.

\subsection{SS-Text: VLMs adaptation at speed-light}
\label{ssec:sstext}

\noindent\textbf{\textit{Constrained linear probing}.} Currently, VLMs are adapted through logistic regression techniques, which combine visual supervisory signals with text knowledge \cite{gao2021clip,yu2023task,lin2023crossmodal,clap24,lp24}. Generally, such methods can be interpreted from a constrained optimization standpoint, in which the new classifier, $\W$, is optimized to minimize a cross-entropy loss and remain close to the class-text prototypes:
\begin{equation}
\label{eq:objective}
    \phantom{.}\min_{\W} \
    \mathcal{L}(\W) = - \frac{1}{N} \sum\limits_{i=1}^{N} \sum\limits_{c=1}^{C}
    y_{ic} \ \text{ln}(p_{ic}(\W)) + \frac{\lambda}{2} \sum\limits_{c=1}^{C} ||\ww_c - \ttt_c||_{2}^{2}.
\end{equation}
Popular choices in the literature involve solving Eq. \ref{eq:objective} via gradient descent, which usually provides satisfactory results since the overall problem is convex. However, the computational effort of computing gradient steps limits its feasibility as a solution in the proposed transductive setting. To alleviate this issue, we propose a novel \textit{training-free} solver to find an approximate solution.

\noindent\textbf{\textit{Training-free approximation}.} We can disentangle Eq. \ref{eq:objective} as the sum of two terms, $\mathcal{L} = g_1 + g_2$, for some $\lambda > 0$, such that:
\begin{align}
   & g_1 = - \frac{1}{N} \sum\limits_{i=1}^{N} \sum\limits_{c=1}^{C}
    y_{ic} \ (\vv^\top \ww_{c}  / \temp) + \frac{\lambda}{2} \sum\limits_{c=1}^{C} ||\ww_c - \ttt_c||_{2}^{2} , \label{g1} \\
   & g_2 = \frac{1}{N} \sum\limits_{i=1}^{N} \text{ln} \left( \sum\limits_{j=1}^{C}
    \text{exp}(\vv^\top \ww_{j}  / \temp) \right), \label{g2}
\end{align}
where $g_1$ controls the hard label assignment, and $g_2$ the soft relationship between categories. We now approximate the solution of Eq.~\ref{eq:objective} by focusing on the first term and minimizing its gradients w.r.t. each $\ww_c$:
\begin{equation}
\label{eq:gradient}
\frac{\partial g_1}{\partial \ww_c} = - \frac{1}{N} \sum\limits_{i=1}^{N} (y_{ic} \ \vv / \temp) + \lambda(\ww_c - \ttt_c).
\end{equation}
Given any $\lambda > 0$, the expression in $g_1$ is convex w.r.t each $\ww_c$, as it is the sum of linear and convex functions. Hence, we can calculate its minimum as:
\begin{equation}
\label{eq:solver}
\ww_c^* = \phantom{.}\arg\min_{\ww_c} \ \frac{\partial g_1}{\partial \ww_c} =  \frac{1}{\lambda N \temp} \sum\limits_{i=1}^{N} y_{ic} \vv + \ttt_c.
\end{equation}
Note that this solution, denoted as SimpleShot-Textual (SS-Text), is a linear combination of textual (\textit{right}) and visual (\textit{left}) prototypes. Hence, we fix $\lambda = 1/(N\temp)$ motivated by: \textit{i)} omitting the effect of $\tau$, and \textit{ii)} diminishing the effect of textual priors as more support data is available.

\section{Experiments}
\label{sec:experiments}

\subsection{Setup}

\noindent\textbf{\textit{Medical vision-language models}.} Three major medical image modalities are employed in this work: histology, fundus, and chest X-ray (CXR) images. \textbf{Histology}: CONCH \cite{CONCH} is used as a histology-specialized model, using a customized larger-scale ViT-B/16 visual backbone. \textbf{Ophtalmology:} FLAIR \cite{FLAIR}, the first VLM for retina imaging, is selected. \textbf{CXR}: We used CONVIRT \cite{convirt} pre-trained on MIMIC \cite{mimic}. FLAIR and CONVIRT follow the same architectural design: the text encoder is BioClinicalBERT \cite{bioclinicalbert}, and the vision encoder is ResNet-50.

\noindent\textbf{\textit{Downstream tasks}.} The selected tasks contain $4$ to $19$ categories and tackle fine-grained tissue/disease classification or grading scenarios. \textbf{Histology}: the datasets involve three different organs and cancer types: colon in NCT-CRC \cite{kather2018100}, prostate Gleason grading in SICAPv2 \cite{silva2020going}, and SkinCancer \cite{kriegsmann2022deep}. \textbf{Ophtalmology}: the popular diabetic retinopathy (DR) grading is included, using MESSIDOR \cite{messidor}. Analogously, myopic maculopathy staging in MMAC \cite{mmac}, and fine-grained disease detection in FIVES \cite{fives}, are assessed. \textbf{CXR}: detection of different diseases from frontal CXR is targeted using popular datasets: a subset of five CheXpert \cite{irvin2019chexpert} categories, employed in \cite{MedCLIP}, 19 fine-grained categories in NIH-LT \cite{nih,nih_lt}, and COVID detection in pneumonia cases \cite{covid1,covid2}.

\noindent\textbf{\textit{Conformal prediction algorithms}.} Three popular non-conformity scores are employed: LAC \cite{lac}, and two adaptive approaches, APS \cite{aps}, and RAPS \cite{raps}, to generate prediction sets at error rates of $\alpha \in \{0.1, 0.05\}$. The hyper-parameters in RAPS are set to $k_{\text{reg}}=1$ and $\lambda=0.001$. These values provided stable performance in \cite{raps}.

\noindent\textbf{\textit{Experimental protocol and metrics}.} The adaptation of medical VLMs assumes the presence of a few-shot adaptation subset, with $K \in \{1, 2, 4, 8, 16\}$. We define a realistic, validation-free setting in which this unique support set adapts the VLM and calibrates conformal prediction methods. It is worth mentioning that this implies that the model selection, i.e., how the adaptation methods are trained, does not rely on validation data. Thus, the training hyper-parameters are constant across tasks. Three conformal prediction settings are explored in the context of VLMs: \textit{i}) \textbf{SCP}, the standard setting studied in vision classifiers \cite{raps,conftr,ding2024class,conftr_24} which operates over zero-shot predictions obtained following Section \ref{ssec:zero-shot} to produce conformal sets, \textit{ii}) \textbf{Adapt+SCP}, which first adjusts an adapter on top of pre-trained features and then follows the SCP over the new output predictions, using the same support set for adaptation and calibration; and \textit{iii}) The proposed \textbf{full conformal adaptation}, which jointly adapts and calibrates in a transductive fashion, as detailed in Section \ref{ssec:fca}. For all these strategies, CP is evaluated when using at least $K \geq 4$, which ensures an appropriate quantile search in Eq. \ref{eq:threshold}. Test data is obtained from each dataset and remains a constant partition, and the support set is retrieved from the training subset. All experiments are repeated 20 times using different random seeds. Class-wise balanced accuracy (ACA) is included to evaluate the discriminative performance, as suggested in \cite{metrics}. Also, figures of merit typically employed in conformal prediction settings are computed: coverage (“Cov.”), average set size (“Size”), and class-conditioned coverage gap (“CCV”) \cite{ding2024class}.

\noindent\textbf{\textit{Adaptation methods}.} SS-Text is compared to recent black-box strategies for adapting VLMs. First, gradient-based trained solutions are used. For vanilla linear probing (LP), we follow ZS-LP in \cite{clap24}. CLAP \cite{clap24} and LP++ \cite{lp24,shakeri2024few} are also integrated, which are trained using full-batch gradient descent for 300 epochs as in \cite{clap24}, using SGD optimizer with momentum of $0.9$. The initial learning rate in LP and CLAP is fixed to $0.1$ as in \cite{clap24}, using a cosine-scheduler decay. For LP++, the learning rate is data-driven, as proposed by the authors. Furthermore, training-free solvers are included: SimpleShot \cite{simpleshot}, as a vision-only method; and vision-language strategies such as TIP-Adapter \cite{zhang2021tip} (with $\alpha_{\text{tip}}$ and $\beta_{\text{tip}}$ fixed to 1.0) and LP++ \cite{lp24,shakeri2024few} initializations. 

\subsection{Conformal prediction results}

This section evaluates the performance of the proposed FCA framework compared to the more traditional SCP or Adapt+SCP. Note that we leverage SS-Text as an adaptation strategy since, in this section, we are only interested in validating the conformalization strategy.

\begin{table}[t!]
\caption{\textbf{Conformal prediction results} using 16 shots, with three popular non-conformity scores, and two error levels, $\alpha \in \{0.05, 0.1\}$. Average across modalities and tasks. “$\downarrow$'' indicates smaller is better. \textcolor{granate}{Red} highlights \textcolor{granate}{unsatisfied error rates}.}
\label{main_results}
\centering
\begin{tabular}{llccccccc}
\toprule
\multicolumn{2}{c}{\multirow{2}{*}{Method}} & & \multicolumn{3}{c}{$\alpha=0.10$} &  \multicolumn{3}{c}{$\alpha=0.05$}   \\ \cmidrule(l){4-6}\cmidrule(l){7-9}  
\multicolumn{2}{c}{}      & ACA$\uparrow$       & Cov.   & Size$\downarrow$ & CCV$\downarrow$ & Cov.   & Size$\downarrow$ & CCV$\downarrow$  \\
\midrule 
\multicolumn{1}{c}{\multirow{3}{*}{\rotatebox{90}{\textbf{LAC}}}} &
SCP                                     & 50.2                        & 0.890      & 3.99                        & 9.96                        & 0.951      & 4.88                       & 5.68 \\
& Adapt+SCP                             & \best{67.1}\imp{+16.9}      & \nc{0.842} & \best{2.40}\imp{-1.59}      & 11.17\wor{+1.21}            & \nc{0.921} & \best{3.07}\imp{-1.81}     & 6.87\wor{+1.19} \\
& \our FCA (\textit{Ours})              & \our \best{67.1}\imp{+16.9} & \our 0.896 & \our 2.91\imp{-1.08}        & \our \best{8.38}\imp{-1.58} & \our 0.952 & \our 3.56\imp{-1.32}       & \our \best{5.02}\imp{-0.66} \\
\midrule 
\multicolumn{1}{c}{\multirow{3}{*}{\rotatebox{90}{\textbf{APS}}}} &
SCP                                     & 50.2                        & 0.900      & 4.05                        & 9.59                        & 0.952      & 4.88                       & 5.54 \\
& Adapt+SCP                             & \best{67.1}\imp{+16.9}      & \nc{0.858} & \best{2.56}\imp{-1.49}      & 8.57\imp{-1.02}             & \nc{0.924} & \best{3.19}\imp{-1.69}     & 6.08\wor{+0.54} \\
& \our FCA (\textit{Ours})              & \our \best{67.1}\imp{+16.9} & \our 0.898 & \our 3.06\imp{-0.99}        & \our \best{6.12}\imp{-3.47} & \our 0.949 & \our 3.67\imp{-1.21}       & \our \best{4.24}\imp{-1.30} \\
\midrule 
\multicolumn{1}{c}{\multirow{3}{*}{\rotatebox{90}{\textbf{RAPS}}}} &
SCP                                     & 50.2                        & 0.901      & 4.16                        & 9.55                        & 0.952      & 5.12                       & 5.57 \\
& Adapt+SCP                             & \best{67.1}\imp{+16.9}      & \nc{0.856} & \best{2.55}\imp{-1.61}      & 8.64\imp{-0.91}             & \nc{0.923} & \best{3.17}\imp{-1.95}     & 6.12\wor{+0.55} \\
& \our FCA (\textit{Ours})              & \our \best{67.1}\imp{+16.9} & \our 0.898 & \our 3.05\imp{-1.11}        & \our \best{6.21}\imp{-3.34} & \our 0.951 & \our 3.66\imp{-1.46}       & \our \best{4.23}\imp{-1.34} \\
\bottomrule
\end{tabular}
\end{table}

\begin{figure*}[t!]
    \begin{center}
        \begin{tabular}{ccc}

         \includegraphics[width=.32\linewidth]{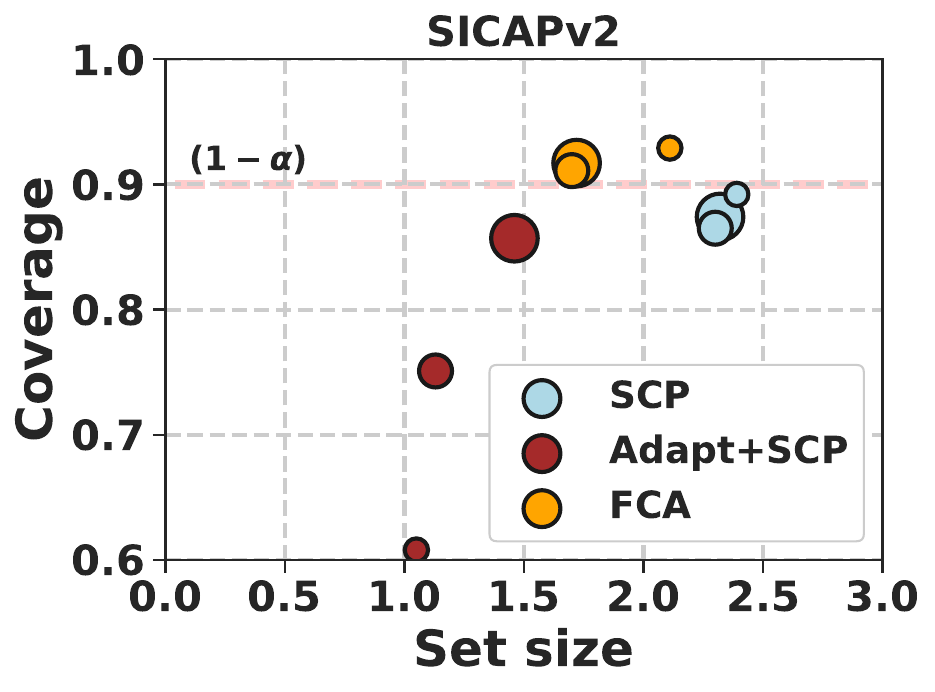} &
         \includegraphics[width=.32\linewidth]{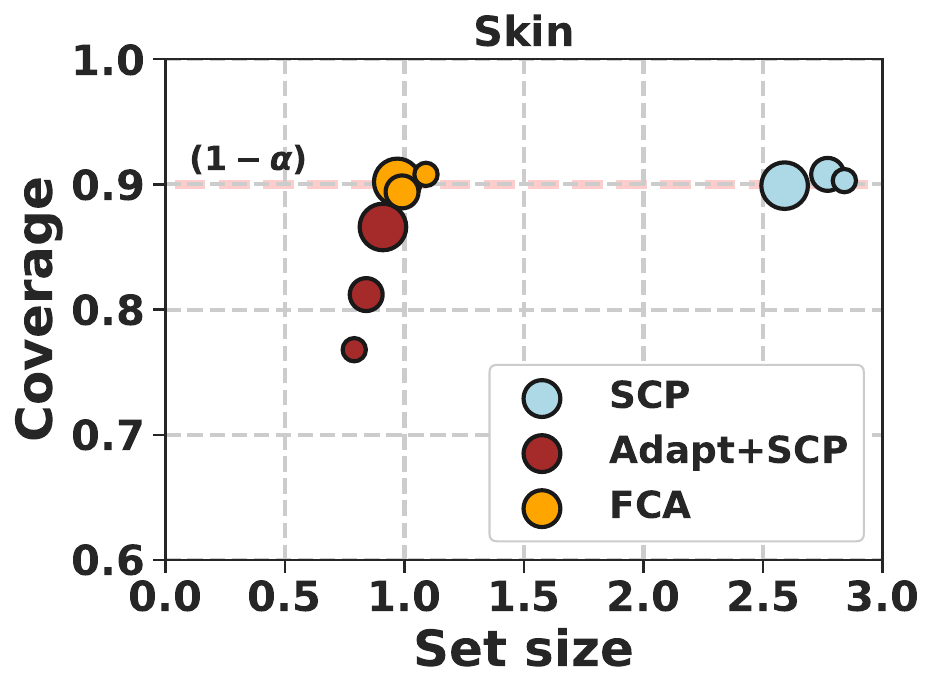} &
         \includegraphics[width=.32\linewidth]{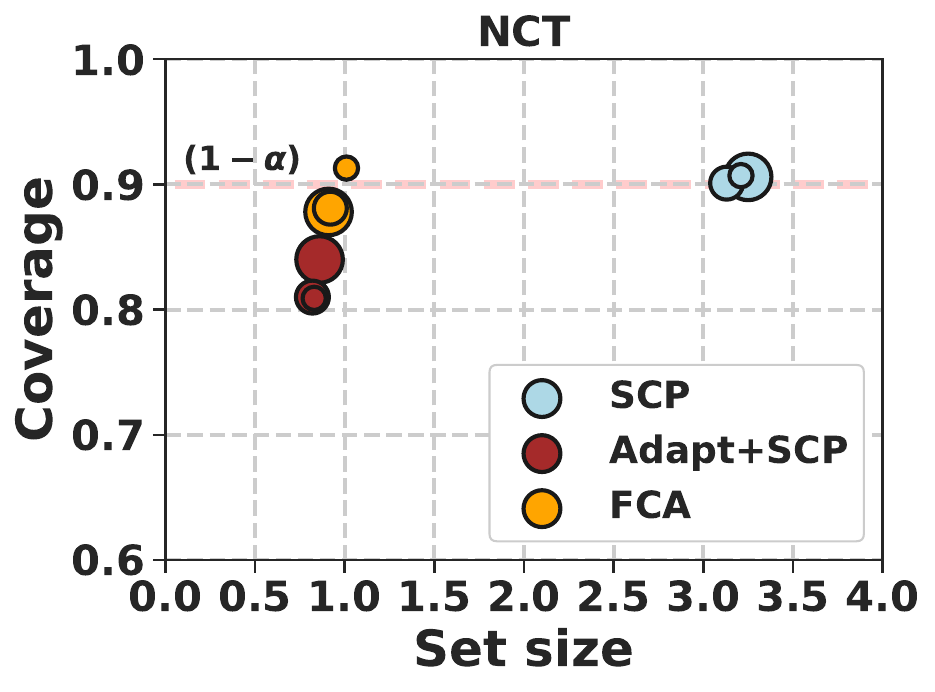} \\

         \includegraphics[width=.32\linewidth]{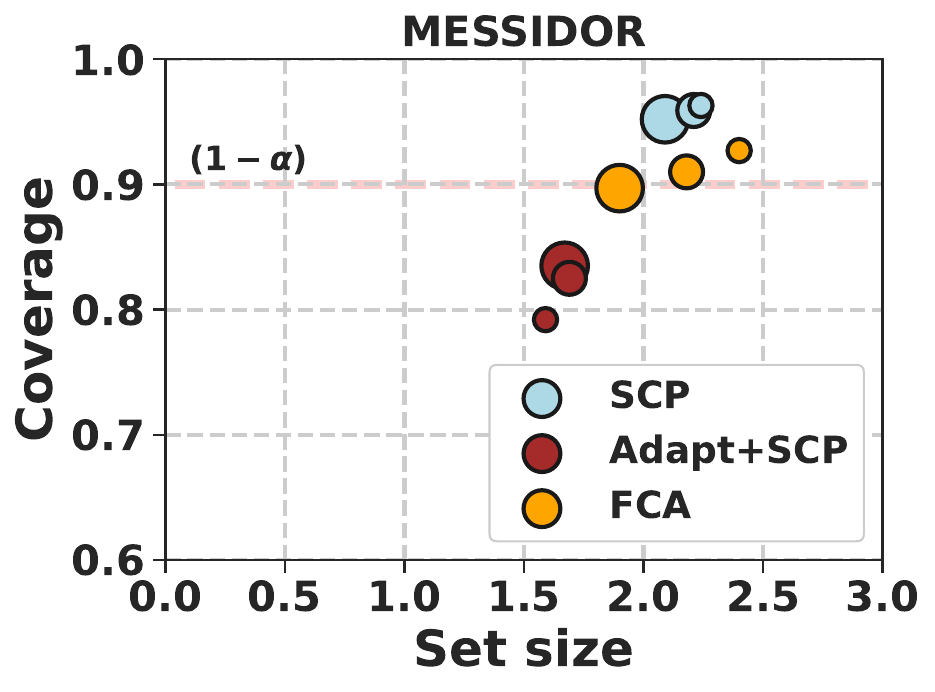} &
         \includegraphics[width=.32\linewidth]{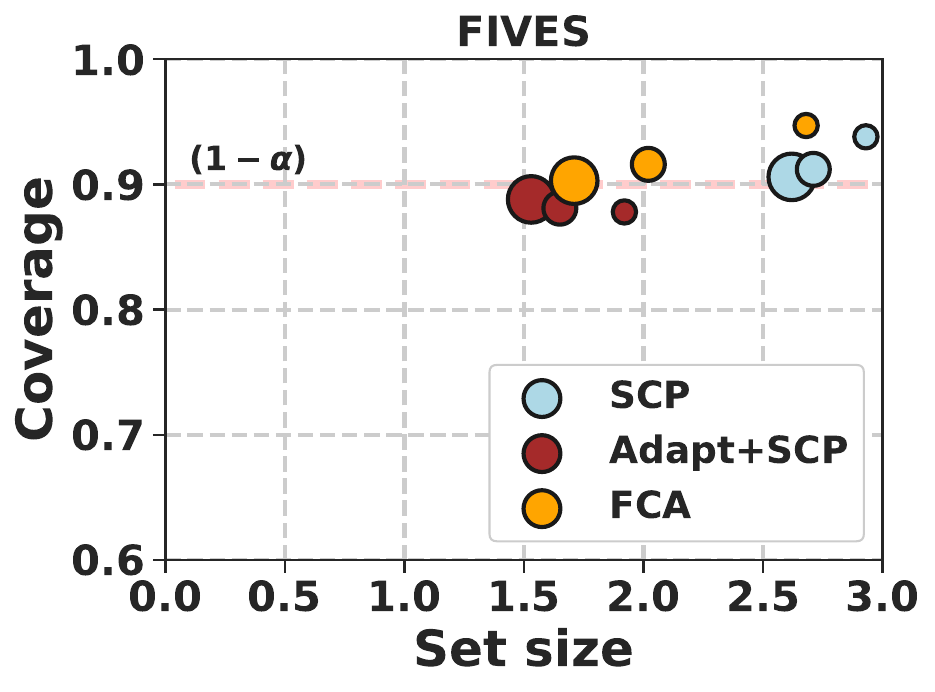} &
         \includegraphics[width=.32\linewidth]{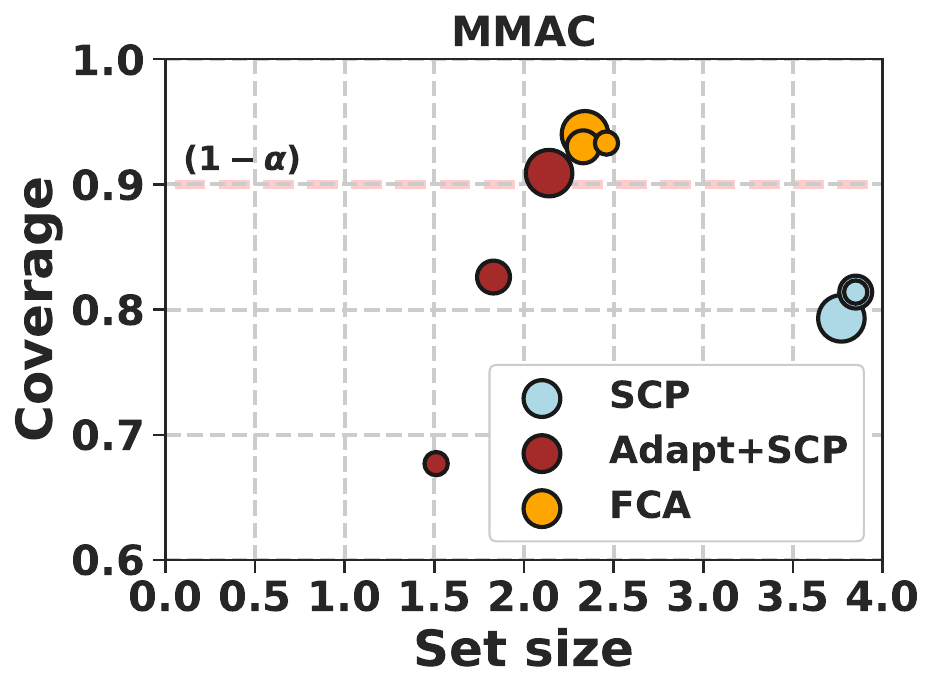} \\

         \includegraphics[width=.32\linewidth]{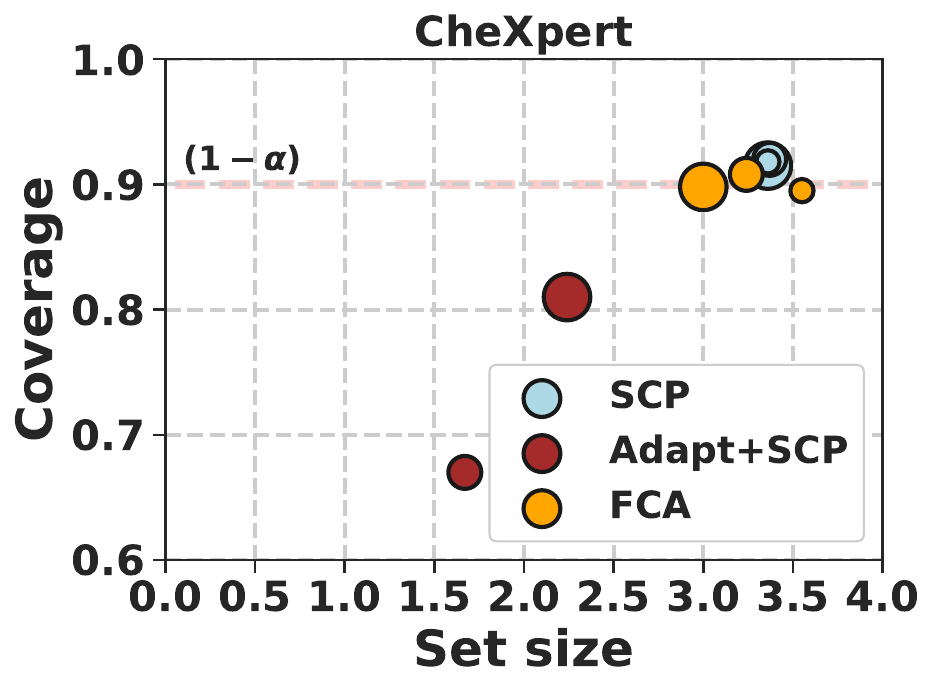} &
         \includegraphics[width=.32\linewidth]{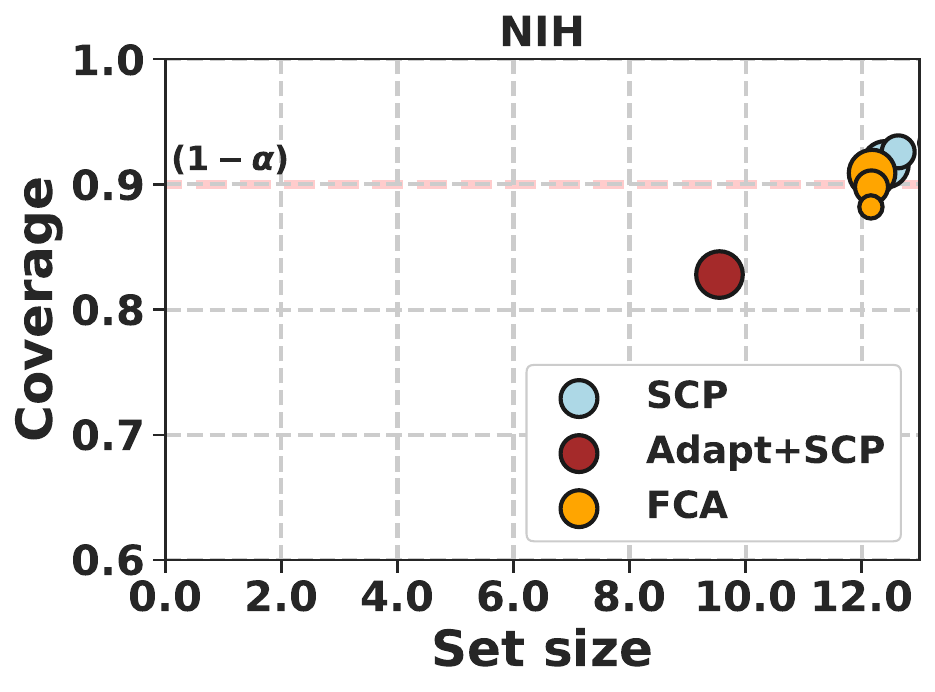} &
         \includegraphics[width=.32\linewidth]{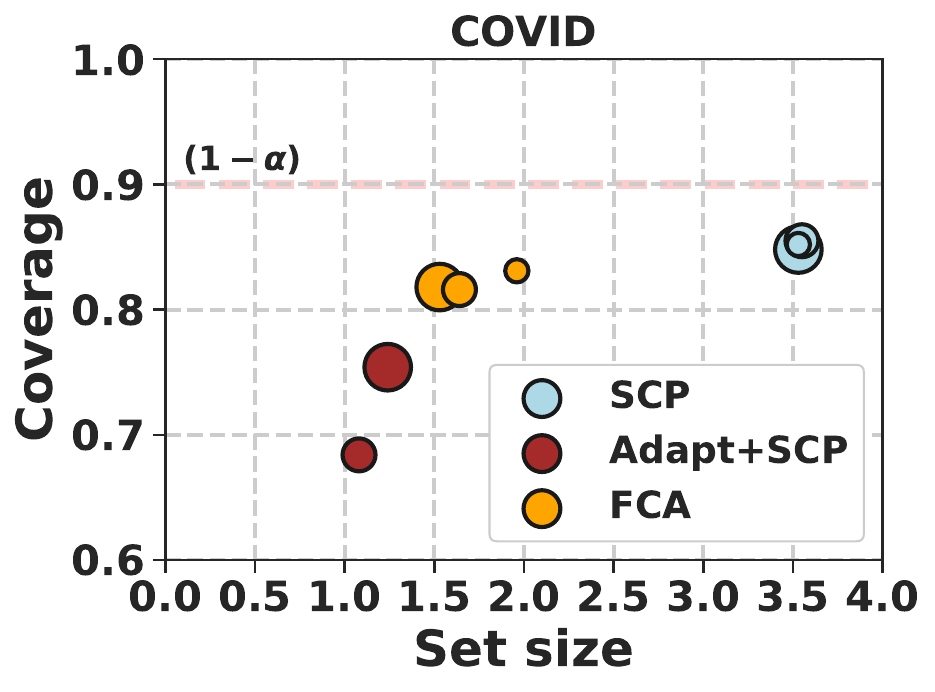} \\

        \end{tabular}
        \caption{\textbf{Conformal prediction results per dataset}. Results were obtained using $\alpha=0.10$, and LAC \cite{lac}. Each dot represents the performance, with increasing size correlated with the number of shots for adaptation, i.e., $K \in \{4, 8, 16\}$.}
        \label{fig:cp_datasets}
    \end{center}
\end{figure*}

\noindent\textbf{\textit{Adapting VLMs with accuracy and reliability}.} Results in Table \ref{main_results} show a near-satisfied average coverage using SCP. However, the results are similar for all non-conformity measures, i.e., LAC, APS, and RAPS. This observation showcases the limited effect of recent efforts devoted to better adaptive scores in the context of zero-shot medical VLMs. When the VLMs are adapted, the accuracy is improved by an average of $+16.9$ points, demonstrating the necessity of transfer learning. Even though Adapt+SCP reduces the produced set sizes, \textit{leveraging the available shots for both adaptation and conformalization naively penalizes coverage}. Thus, its produced sets are unreliable. In contrast, the proposed FCA excels in both aspects. First, it provides discriminative performance on par with the standard adaptation. Second, it keeps the marginal coverage at the desired error level, reducing the produced sets by nearly $27\%$ compared to SCP. FCA also improves the more fine-grained class-conditional coverage, decreasing its violation up to $\sim15\%$ for LAC and $\sim35\%$ for adaptive scores.

\noindent\textbf{\textit{Detailed, per-dataset findings}.} When observing the results in Figure \ref{fig:cp_datasets}, one can notice that exact marginal coverage is usually not achieved — not even for SCP, e.g., in SICAPv2 or MMAC. In contrast to the main corpus of conformal prediction evaluating CP in vision, medical image classification poses more challenging scenarios to ensure the exchangeability principle. First, testing data comes from patients who are naturally inaccessible during calibration. Second, the label-marginal distribution of the testing data is unknown, and typical few-shot adaptation pipelines assume a balanced support set. Combined with the finite sampling error, these facts explain such observation and showcase the necessity of developing CP techniques tailored to medical image analysis. However, it is worth mentioning that the proposed FCA framework consistently improves average coverage satisfaction (especially when compared with the naive Adapt+SCP strategy) while improving the prediction sets by large margins.

\subsection{Few-shot adaptation benchmark}

This section explores the performance, in the discriminative aspect, of the proposed SS-Text solver compared to SoTA linear probing strategies. We also evaluate the feasibility of each method to serve as a solver for full conformal adaptation, which refers to its efficiency, measured in terms of how many fits (evaluating one category for a query image) they perform per second in such a setting.

\begin{figure*}[t!]
    \begin{center}
        \begin{tabular}{cc}

         \includegraphics[width=.50\linewidth]{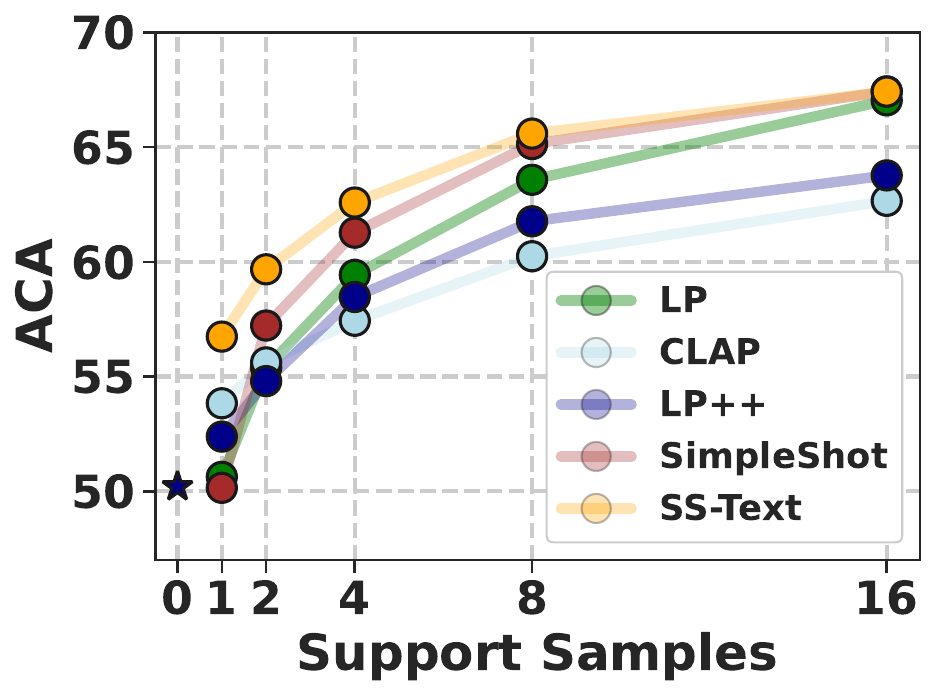} & 
         \includegraphics[width=.50\linewidth]{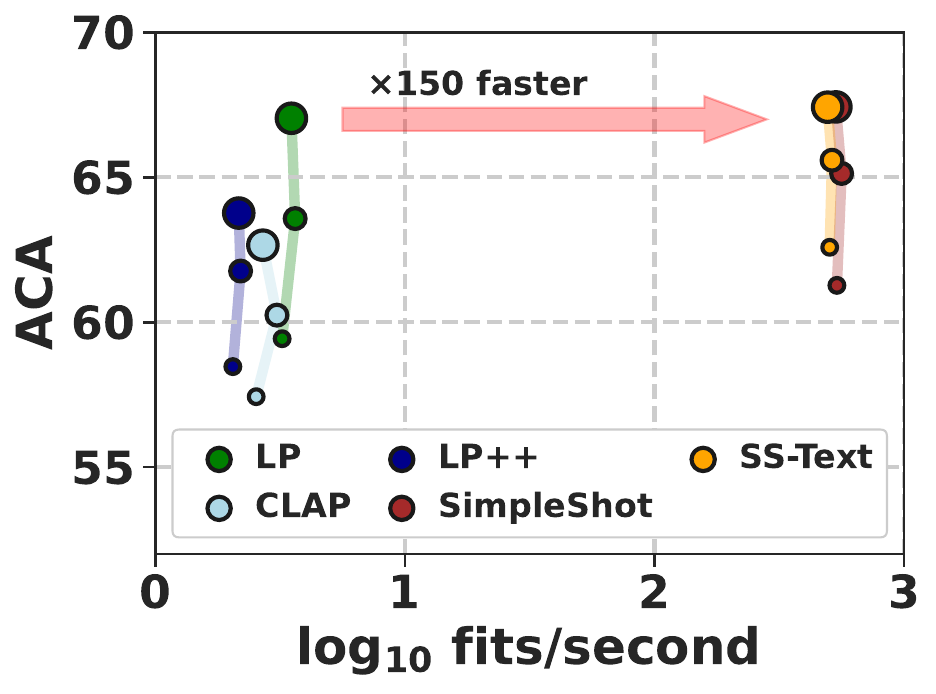} \\
         \textbf{(a) Few-shot adaptation} & \textbf{(b) FCA Efficiency} \\

        \end{tabular}
        \caption{\textbf{Few-shot adaptation performance (a) and efficiency analysis (b)}. Efficiency, regarding full conformal adaptation predictions: images per class - per second. The dot size in (b) indicates the number of shots, i.e., $K \in \{4, 8, 16\}$.}
        \label{fig:accuracy_efficiency}
    \end{center}
\end{figure*}

\noindent\textbf{\textit{Discriminative performance}.} Figure \ref{fig:accuracy_efficiency}(a) demonstrates that the proposed training-free SS-Text is a competitive solution in terms of accuracy. This is especially the case for the low-shot scenario, in which SS-Text outperforms popular recent solutions, such as CLAP \cite{clap24} or LP++ \cite{lp24}. Also, SS-Text is robust when the number of support samples increases, where basic linear probe baselines provide competitive results. We argue that these results are the product of the explored realistic setting, in which no validation data is accessible for defining a fine-grained scheduler and early-stopping intensive gradient-based techniques. When trained on a fixed scheduler, methods such as CLAP or LP++ might suffer in finding a homogeneous optimal configuration for all datasets.

\noindent\textbf{\textit{Efficiency}.} Figure \ref{fig:accuracy_efficiency}(b) showcases the extreme efficiency of SS-Text when performing sample-wise predictions within the proposed full conformal adaptation setting, being approximately $150\times$ faster than gradient-based approaches. For example, if using a gradient-based linear probe strategy (LP), the inference of each image in the Skin dataset ($M=30,000$; $C=16$) would require at least $2.8$ seconds per image (in contrast to $30$ milliseconds with SS-Text), which accumulates approximately $23$ hours for the whole dataset, thus being infeasible its deployment in FCA. These figures underscore the necessity of efficient, training-free solvers to provide trustworthy sets in real-world scenarios.

\subsection{In-depth studies}

\noindent\textbf{\textit{Role of $\lambda$}.} We fixed $\lambda=1/(N\temp)$ in Eq.~\ref{eq:solver}, assuming that the more supervision signals we leverage for adaptation, the less effect of the constraint is desired. Table \ref{lambda} assesses this adaptive model selection, compared to fixing it for different values, concretely $\lambda \in \{0.1, 1.0, 10\}$. The proposed adaptive strategy achieves the best performance balance between low-shot (e.g., $K \in \{1, 2, 4\}$), and a larger number of supervisory signals (e.g., $K \in \{8, 16\}$). This favorable comparison also extends to the strategy proposed in CLAP \cite{clap24}, where the authors proposed a class-wise multiplier based on zero-shot performance on the support set. 

\noindent\textbf{\textit{Comparison to training-free baselines}.} SS-Text is also compared to popular training-free baselines in Table \ref{lambda} (\textit{top}). The proposed method consistently outperforms TIP-Adapter \cite{zhang2021tip} and LP++ (training-free version) \cite{lp24}. Also, compared to SimpleShot \cite{simpleshot}, SS-Text shows favorable performance. Note that SimpleShot does not rely on any text information, and thus, its performance on the low-shot regime is hampered. Such results showcase the importance of integrating text supervision for the few-shot adaptation of medical VLMs. Note that the performance gap between SimpleShot and SS-Text decreases with the amount of data since SS-Text naturally approximates SimpleShot with large $N$ values.

\begin{figure*}[t!]
    \begin{center}
        \begin{tabular}{cccc}

         \includegraphics[width=.24\linewidth]{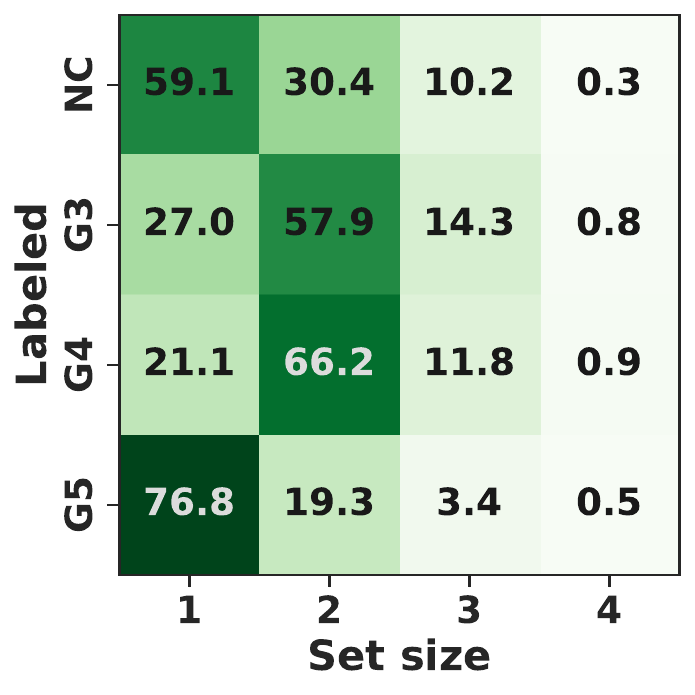} &
         \includegraphics[width=.24\linewidth]{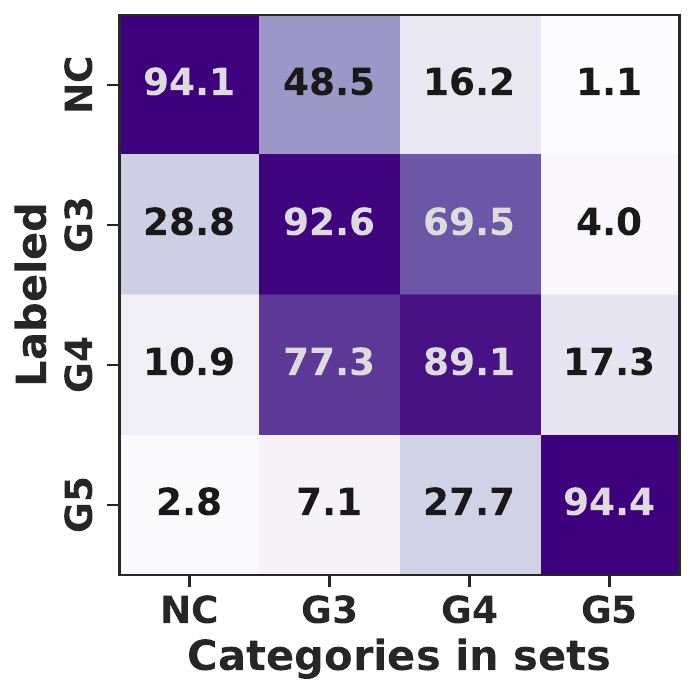} &
         \includegraphics[width=.24\linewidth]{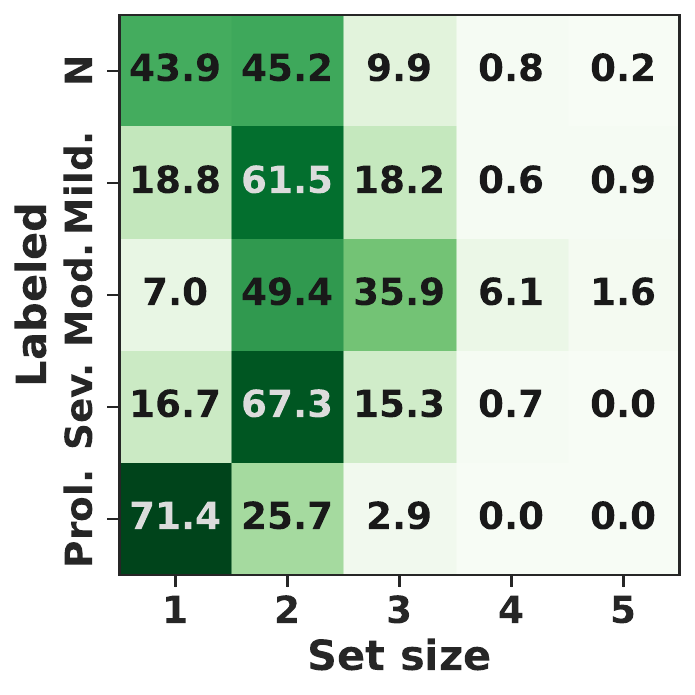} &
         \includegraphics[width=.24\linewidth]{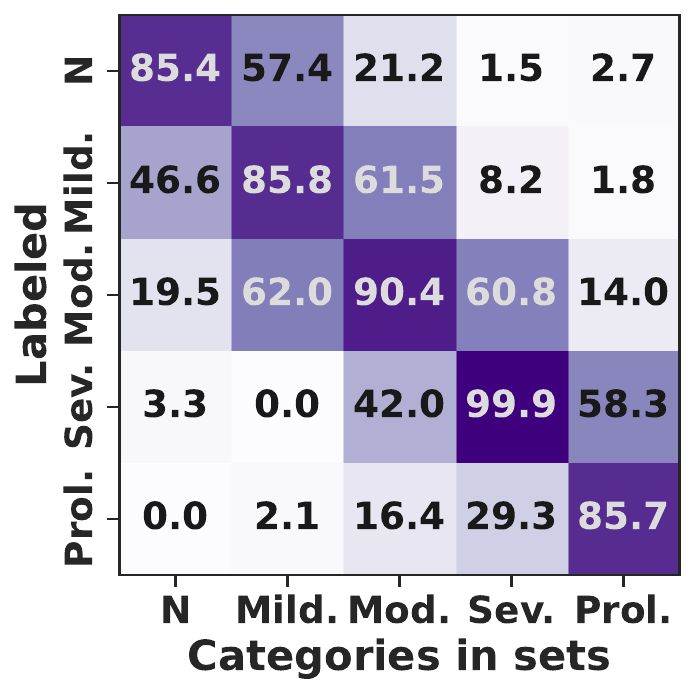} \\

         \multicolumn{2}{c}{\textbf{(a) SICAPv2}} &  \multicolumn{2}{c}{\textbf{(b) MESSIDOR}} \\

        \end{tabular}
        \caption{\textbf{Qualitative evaluation} of conformal prediction for grading tasks, i.e., Gleason for histology (a) and diabetic retinopathy for retina (b). \textbf{\textcolor{drakgreen}{Left}}: set size distribution per class. \textbf{\textcolor{drakpurple}{Right}}: label frequency in sets corresponding for each category. Results using FCA with $K=16$, SS-Text solver, and $\alpha=0.1$.}
        \label{fig:qualitative}
    \end{center}
\end{figure*}

\noindent\textbf{\textit{Qualitative assessment}.} We now depict some figures highlighting the utility of conformal prediction for medical image analysis. Concretely, we focus on two grading tasks for which well-known inter-observer variability exists. Figure \ref{fig:qualitative} introduces the set size distribution and the frequency of class appearance within the predicted sets for each category. First, one can notice that smaller sets are usually retrieved for the more extreme and "easy'' categories, while they tend to increase in categories that present similar patterns. Also, the occurrence matrices unveil an overlap between the latter classes, similar to typically observed inter-rater variability for these tasks \cite{zurich,Galdran2020}. It is worth remembering that such sets present empirical guarantees of giving the correct category for $90\%$ of the cases for tasks with relatively smaller accuracy, i.e., $69\%$ and $60\%$. Thus, the conformal prediction setting smoothly provides information regarding uncertainties natural from grading tasks.

\begin{table}[t!]
\caption{\textbf{Study on how to fix $\lambda$} in SS-Text solver (\textit{bottom}) and comparison with training-free baselines (\textit{top}). Average accuracy across nine datasets.}
\label{lambda}
\centering
\begin{tabular}{llccccc}
\toprule
\multicolumn{1}{c}{\multirow{1}{*}{Method}} & \multicolumn{1}{c}{\multirow{1}{*}{Setting}} & $K=1$     & $K=2$    & $K=4$    & $K=8$    & $K=16$   \\
\midrule
Zero-shot \cite{radford2021learning}   & (only text)              & 50.2 & 50.2 & 50.2 & 50.2 & 50.2   \\
SimpleShot \cite{simpleshot}           & (only vision)            & 50.1 & 57.2 & 61.3 & 65.1 & \textbf{67.4}   \\
TIP-Adapter \cite{zhang2021tip}        & (training-free)          & 55.5 & 55.5 & 60.2 & 62.2 & 63.2   \\
LP++ \cite{lp24}                       & (training-free)          & 50.7 & 51.0 & 51.4 & 52.1 & 53.2   \\
\midrule
SS-Text                                & Fixed $\lambda=0.1/\temp$          & 55.1 & 59.0 & 61.5 & 63.6 & 64.3   \\
SS-Text                                & Fixed $\lambda=1.0/\temp$          & 53.5 & 54.1 & 54.5 & 54.7 & 54.6   \\
SS-Text                                & Fixed $\lambda=10/\temp$           & 51.2 & 51.2 & 51.2 & 51.1 & 51.2   \\
SS-Text                                & $\lambda_c\simeq$ zero-shot perf. \cite{clap24}   & 51.4 & 58.4 & 62.6 & \textbf{65.7} & \textbf{67.4}  \\
\rowcolor{Gray}SS-Text (\textit{Ours}) & $\lambda=1/(N\temp)$                              & \textbf{56.7} & \textbf{59.7} & \textbf{62.6} & 65.6 & \textbf{67.4}   \\
\bottomrule
\end{tabular}
\end{table}

\section{Discussion}
\label{sec:conclusion}

While the proposed FCA has proven remarkable properties compared to SCP, it also presents some limitations. First, it provides theoretical guarantees under exchangeability, which might be unrealistic to achieve in certain medical image analysis scenarios. Second, the framework requires one linear probe fit for each test image and category, which might be prohibitive in dense prediction tasks or when considering many classes. However, the latter scenario is least common in medical image tasks, where only tenths of tags are usually addressed at a time.

\newpage
\begin{credits}
\subsubsection{\ackname} This work was funded by the Natural Sciences and Engineering Research Council of Canada (NSERC) and \textit{Fonds de recherche du Québec} (FRQ). We thank Calcul Québec and Compute Canada. We gratefully acknowledge the DATAIA program for supporting Jose Dolz as a visiting professor at Université Paris-Saclay.
\subsubsection{\discintname} The authors have no competing interests to declare that are relevant to the content of this article.
\end{credits}

\bibliographystyle{splncs04}
\bibliography{refs}

\end{document}